\documentclass[format=sigconf, anonymous=false]{acmart}
\usepackage{subfigure}
\usepackage{enumerate}
\usepackage{graphicx}
\usepackage{multirow}
\usepackage{epstopdf}

\renewcommand\footnotetextcopyrightpermission[1]{}

\setcopyright{acmcopyright}
\copyrightyear{2021}
\acmYear{2021}
\acmDOI{10.1145/1122445.1122456}

\acmConference[ACM Multimedia '29]{Chengdu '29: ACM Multimedia}{October 20–24, 2021,}{Chengdu, China}
\acmBooktitle{Chengdu '29: ACM Multimedia,
	October 20--24, 2021,  Chengdu, China}
\acmPrice{15.00}
\acmISBN{978-1-4503-XXXX-X/18/06}

\acmSubmissionID{2657}

\begin{document}

\title{UIF: An Objective Quality Assessment for Underwater Image Enhancement}

    
%
\author{Yannan Zheng}
\affiliation{%
  \institution{College of Physics and Information Engineering, Fuzhou University}
  \state{Fujian}
  \country{China}}
\email{N191127059@fzu.edu.cn}

\author{Weiling Chen}
\affiliation{%
  \institution{College of Physics and Information Engineering, Fuzhou University}
  \state{Fujian}
  \country{China}}
\email{weiling.chen@fzu.edu.cn}
\author{Rongfu Lin}
\affiliation{%
  \institution{College of Physics and Information Engineering, Fuzhou University}
  \state{Fujian}
  \country{China}}
\email{N191127021@fzu.edu.cn}
%
%
\author{Tiesong Zhao}
\affiliation{%
  \institution{College of Physics and Information Engineering, Fuzhou University}
  \state{Fujian}
  \country{China}}
\email{t.zhao@fzu.edu.cn}
%
\renewcommand{\shortauthors}{Lin, et al.}


\begin{abstract}
Due to complex and volatile lighting environment, underwater imaging can be readily impaired by light scattering, warping, and noises. To improve the visual quality, Underwater Image Enhancement (UIE) techniques have been widely studied. Recent efforts have also been contributed to evaluate and compare the UIE performances with subjective and objective methods. However, the subjective evaluation is time-consuming and uneconomic for all images, while existing objective methods have limited capabilities for the newly-developed UIE approaches based on deep learning. To fill this gap, we propose an Underwater Image Fidelity (UIF) metric for objective evaluation of enhanced underwater images. By exploiting the statistical features of these images, we present to extract naturalness-related, sharpness-related, and structure-related features. Among them, the naturalness-related and sharpness-related features evaluate visual improvement of enhanced images; the structure-related feature indicates structural similarity between images before and after UIE. Then, we employ support vector regression to fuse the above three features into a final UIF metric. In addition, we have also established a large-scale UIE database with subjective scores, namely Underwater Image Enhancement Database (UIED), which is utilized as a benchmark to compare all objective metrics. Experimental results confirm that the proposed UIF outperforms a variety of underwater and general-purpose image quality metrics.

\end{abstract}




\maketitle

\section{Introduction}
\begin{figure}[]
\centering
	\subfigure [Original image]{
	\includegraphics[height=1.5cm,width=1.875cm]{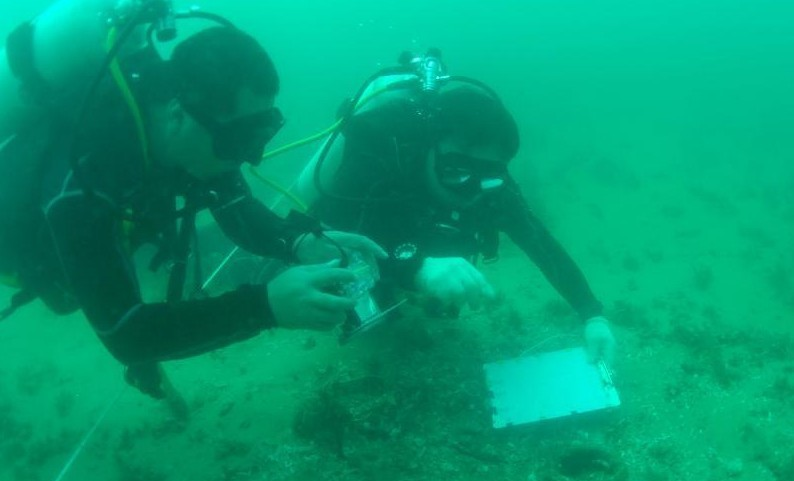}	}
	\subfigure [0.7541/1.7745]{
	\includegraphics[height=1.5cm,width=2cm]{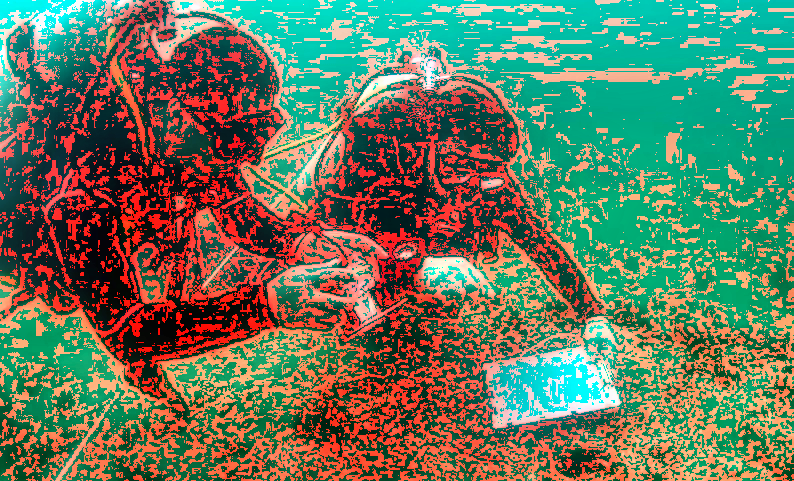}	}
	\subfigure [0.6326/1.5410]{
	\includegraphics[height=1.5cm,width=1.875cm]{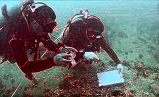}	}
	\subfigure [0.6254/1.4213]{
	\includegraphics[height=1.5cm,width=1.875cm]{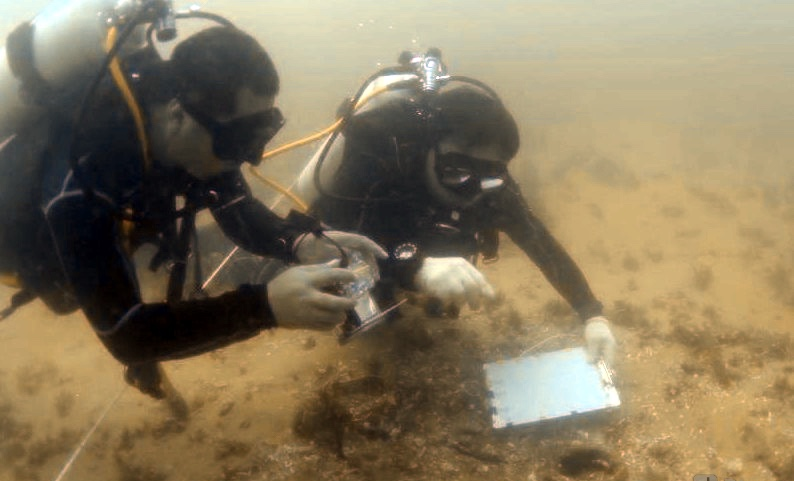}	}\\
	\vspace{-0.1cm}
	\subfigure [Original image]{
	\includegraphics[height=1.5cm,width=1.875cm]{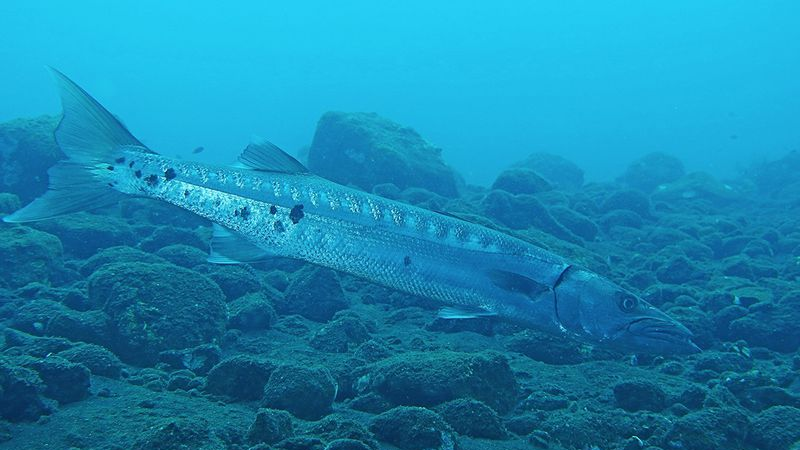}	}
	\subfigure [0.6321/1.6217]{
	\includegraphics[height=1.5cm,width=1.875cm]{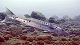}	}
	\subfigure [0.6441/1.6745]{
	\includegraphics[height=1.5cm,width=1.875cm]{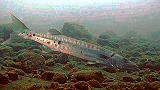}	}
	\subfigure [0.5865/1.5233]{
	\includegraphics[height=1.5cm,width=1.875cm]{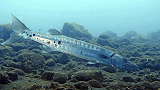}	}\\
	\vspace{-0.1cm}
	\subfigure [Original image]{
	\includegraphics[height=1.5cm,width=1.875cm]{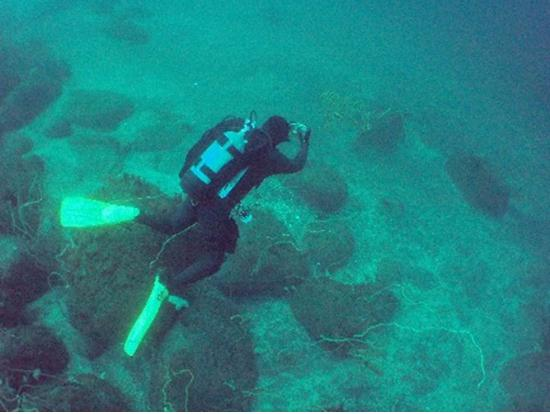}	}
	\subfigure [0.6458/1.6245]{
	\includegraphics[height=1.5cm,width=1.875cm]{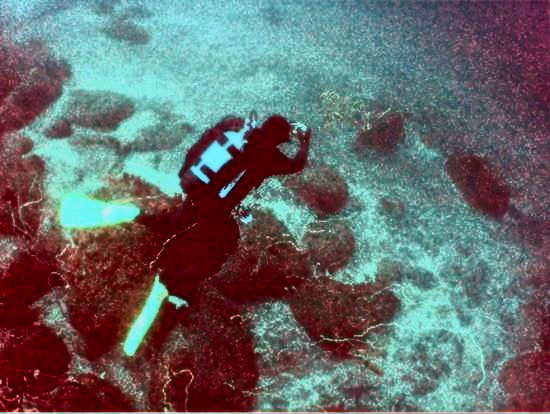}	}
	\subfigure [0.6605/1.7614]{
	\includegraphics[height=1.5cm,width=1.875cm]{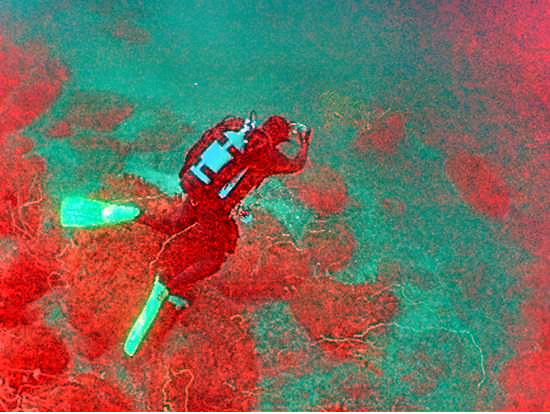}	}
	\subfigure [0.6139/1.4705]{
	\includegraphics[height=1.5cm,width=1.875cm]{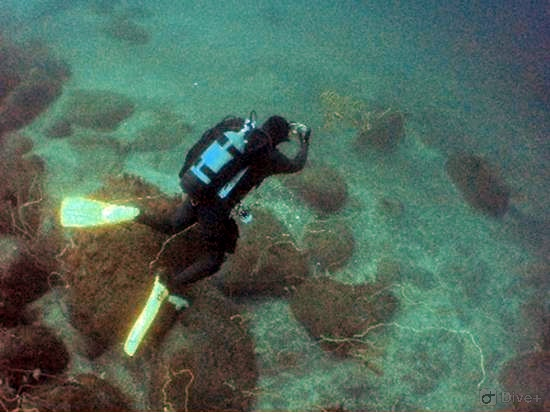}	}\\
	\vspace{-0.1cm}
	\caption{\small Typical underwater images (the first left column) and their scored enhancements (the other three columns), where the scores of each enhanced image are respectively given by UCIQE and UIQM. How to compare the visual quality of enhanced underwater images is still a challenging task.}
	\label{fig:1}
\end{figure}

The underwater optical images bring additional information beyond sonar imaging. However, the complex waterbody and poor light conditions impair the visual quality of underwater images. In practice, Underwater Image Enhancement (UIE) technique is thus necessary to transfer low-quality underwater images to high-quality ones. As shown in Fig. \ref{fig:1}, diversified enhancements are employed in original underwater images, resulting in pictures with higher visual qualities. To compare the UIE algorithms and select optimal results, it is imperative to score these enhanced images, which is still a challenging task.

Until now, the visual quality of images can be evaluated by Image Quality Assessment (IQA), whose popular works include subjective and objective methods. For most of images, human is the ultimate receiver, thus the subjective evaluation is considered to be the most accurate and reliable way of IQA. Recently, ITU has promoted several methods to subjectively evaluate image quality \cite{Recommendation2012}. However, the subjective evaluation also has significant drawbacks: high complexity, high cost and unable to be embedded into real-world systems. As a result, the subjective methods are usually utilized to set benchmark to evaluate objective metrics, and the objective metrics with higher correlations to subjective scores are embedded into real-world systems.  

Existing objective IQA can be classified into full-reference, reduced-reference and no-reference algorithms, subject to the accessibility of ideally unimpaired references. In underwater imaging, an ideally unimpaired image is unable to be obtained, thus the typical reference-based IQA approaches, such as Structural Similarity Index (SSIM) \cite{Wang2020} and Feature Similarity Index (FSIM) \cite{Zhang2017}, are not applicable. On the other hand, no-reference IQA approaches have achieved significant performances to evaluate generic images \cite{Anish2012, Anish2013, Gu2015, Zhang2015, Yang2016}. However, these no-reference approaches are usually designed based on Natural Scene Statistics (NSS), which performs different in underwater environment. Thus, they also fail to evaluate the quality of underwater images.

The Underwater Color Image Quality Evaluation (UCIQE) \cite{yang2015} and Underwater Image Quality Measure (UIQM) \cite{K2016} metrics have been widely used to qualify the enhancement performance of underwater images \cite{odm2016, chang2019, Peng2017, Li2019}. Among them, the UCIQE metric quantifies non-uniform color casts, blurring, and noise in underwater images, and then combines them in a linear manner. The UIQM metric evaluates underwater images with color, sharpness and contrast. These two methods have been proved to achieved good performances in most underwater scenarios at that time. However, this decade has witnessed a booming of UIE algorithms, especially deep-learning-based UIE algorithms. In such case, the UIE may generate complicated color and structure changes which are unable to be evaluated by conventional metrics. From Fig. 1, a higher UCIQE/UIQM score may not represent a better visual quality. For example, some images have obvious reddish color shifts and artifacts as shown in Fig 1. (b), (g) and (k), but these images obtain better UCIQE and UIQM scores. Therefore, it is highly desirable to design an effective IQA metric for objective UIE evaluation.


To address this issue, we exploit the characteristics of underwater images to obtain three types of features: naturalness-related, sharpness-related and structure-related features, which are demonstrated to be effective to assess the underwater image quality. The Support Vector Regression (SVR) are then utilized to fuse all features into a final metric. In addition, we also develop the largest-ever UIE quality database with subjective scores, which helps to guarantee the generalization performance of our model. It can also serve as a benchmark to compare subjective IQA metrics for UIE. In summary, the main contributions of this paper include:
\begin{enumerate}[1)]
\item A large-scale Underwater Image Enhancement Database (UIED), which is the first-of-its-kind database with human-labeled quality scores. The database includes different underwater scenes that are enhanced by 10 representative UIE approaches. The UIED can be utilized as a benchmark to develop and evaluate subjective methods of underwater image quality assessment. 
\item Three types of image features to evaluate the underwater image fidelity in three dimensions. An optimal UIE should generate natural images with sharp textures and high structure similarity to original image. Inspired by this, we propose to extract and model the three types of features based on statistics of underwater images. The effectiveness of features has been proved in ablation study.
\item An SVR-based fusion of features to obtain our Underwater Image Fidelity (UIF) metric. We employ the popular SVR models for feature pooling and regression. To ensure the generalization performance of our UIF metric, it is trained and examined by k-fold validation in UIED. Experimental results reveal the efficiency of our method.
\end{enumerate}

\section{RELATED WORK}
In this section, we review related works of our paper. Among them, the UIE methods are utilized to construct our UIED database. The conventional IQA methods and Underwater IQA methods are compared in our experiments.

\subsection{Underwater Image Enhancement Methods}
In recent years, many UIE methods have been proposed. The existing UIE methods can be divided into three categories: Non-physical-model-based methods \cite{Ancuti2012}, \cite{Fu2014}, \cite{Fu2017}, physical-model based \cite{Drews2016}, \cite{Peng2018}, \cite{Peng2017}, \cite{odm2016} and deep-learning-based methods\cite{Li2019}, \cite{Li2020}. 

Non-physical-model-based methods aim to adjust input image pixel values to
improve visual quality. Ancuti \emph{et al.}  \cite{Ancuti2012} proposed a UIE algorithm with multi-scale fusion strategy. Fu \emph{et al.} \cite{Fu2014} proposed a retinex-based UIE method, which decomposes the reflectance and illumination of underwater images. In \cite{Fu2017}, they also proposed a two-step enhancement procedure, which includes a color correction and a contrast enhancement.

Physical-model-based methods construct physical models for underwater  degradation. Li \emph{et al.} \cite{odm2016} proposed a contrast enhancement algorithm which combined with an image dehazing model. In \cite{Drews2016}, an Underwater Dark Channel Prior (UDCP) was proposed based on the fact that the information of red channel is undependable. Peng \emph{et al.} \cite{Peng2018} proposed a Generalized Dark Channel Prior (GDCP) that incorporates adaptive color correction into an image formation model for UIE. Based on image blurriness and light absorption, they also \cite{Peng2017} proposed a depth estimation method for underwater scenes.

Deep-learning-based methods have led a fast development and offered state-of-the-art performance in many UIE tasks. Li \emph{et al.} \cite{Li2019} proposed an end-to-end deep network, namely UWCNN, to address the UIE problem for diverse underwater images. In \cite{Li2020}, they also designed a Water-Net model which is trained on paired underwater images and the corresponding reference images.

In addition, the commercial application dive+ \cite{dive+} has also been released in iTunes store, with good performances of underwater enhancement.

\subsection{Conventional IQA Methods}
IQA plays an important role in many computer vision problems. In recent years, metrics have been used to evaluate enhancement or restoration performance for underwater images. 

Most traditional no-reference IQA metrics are based on NSS regularities. Anish \emph{et al.} extracted effective statistical features to evaluate image quality \cite{Anish2012}. The Natural Image Quality Evaluator (NIQE) model proposed by Mittal \emph{et al.} \cite{Anish2013} extracted a set of local features from an image. Gu \emph{et al.} \cite{Gu2015} designed No-reference Free Energy Robust Metric (NFERM) by adding features of Human Vision System (HVS). Gu \emph{et al.} \cite{GUGU2015} solved the no-reference IQA for blur images using a sharpness metric in autoregressive parameter space. Xue \emph{et al}. \cite{Xue2014} combined gradient magnitude with laplacian of gaussian to predict image quality. Kang \emph{et al.} \cite{cnniqa} investigated a Convolutional Neural Network (CNN) to jointly learn features for IQA. However, an underwater image is always degraded by light absorption and scattering, which may not be captured by these NSS models effectively. Thus, these NSS-based approaches are not effective enough for underwater images. 

Meanwhile, some no-reference IQA methods have been proposed for enhanced image quality assessment. Fang \emph{et al.} \cite{fang2015} proposed a blind contrast quality metric based on image skewness, kurtosis, and entropy. Gu \emph{et al.} \cite{Gu2018} proposed a blind metric according to 17 features through analysis of images. These methods may not perform well in underwater IQA due to lack of consideration on the underwater imaging models and image features. Liu \emph{et al.} \cite{2019Enhanced} used synthetic data to generate the ideal reference image, thereby converting the evaluation of enhanced images into a full-reference IQA problem. However, its synthetic data only cover very limited types of underwater conditions, which may be insufficient to simulate underwater image distortions. Thus, these methods are also not reliable to apply for underwater image enhancement.

\subsection{Underwater IQA Methods}
Until now, there are two underwater IQA methods that are widely applied to evaluate UIE approaches. They are the UCIQE \cite{yang2015} and UIQM \cite{K2016} metrics.

Yang \emph{et al.} designed the UCIQE metric in \cite{yang2015}. The UCIQE transforms an underwater image from RGB color space to CIELab color space, which is more consistent with HVS. It quantifies non-uniform color casts, blurring and noise in underwater monitor images and then linearly combines these three components. A higher UCIQE score indicates the result has a better balance among the chroma, saturation and contrast.

Panetta \emph{et al.} proposed the UIQM inspired by HVS in \cite{K2016}. The UIQM comprises a colorfulness measure, asharpness measure, and a contrast measure for underwater images, based on the underwater image modeling presented in \cite{K2016}. The choice of weighted coefficients depends on the application purpose. A higher UIQM score denotes the result is more consistent with human visual perception. 

As shown in Fig. 1, the two metrics also have lower correlations to human scores when the enhancement distortion is complex. However, due to the wide applications of deep learning models, the UIE technique has inevitably brought complex warping and color changes. To model the impacts of these complex distortions, we suggest to utilize a deep learning network. This also requires a large-scale database to train and test the deep-learning-based model.


\section{PROPOSED UIED DATABASE}
There is a lack of publicly available large-scale UIE database with human subjective scores. To fill this void, the UIED dataset is developed as the largest-ever subjective database of UIE images, which can serve as a benchmark to develop and evaluate objective approaches. In this section, we will elaborate the preparation of images, subjective test and post-processing to construct this database.

\subsection{Preparation of Enhanced Underwater Images}
\begin{figure}[htbp]
\centering
\subfigure [Coral] {
\includegraphics[height=2cm,width=2.5cm]{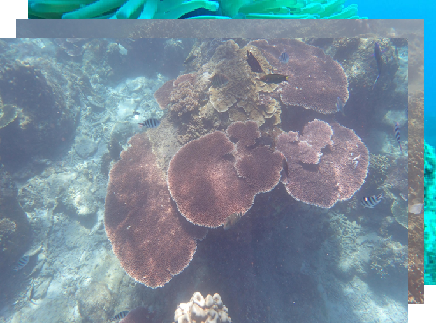}}
\subfigure [Marine life] {
\includegraphics[height=2cm,width=2.5cm]{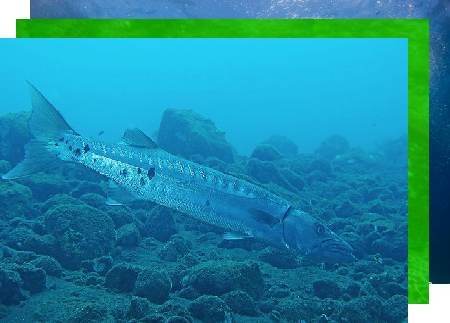}}
\subfigure [Seabed rock] {
\includegraphics[height=2cm,width=2.5cm]{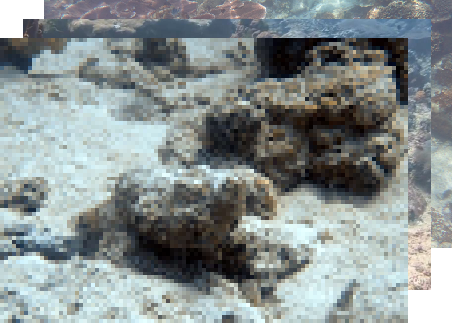}}\\
\vspace{-0.1cm}
\centering
\subfigure [Sculpture] {
\includegraphics[height=2cm,width=2.5cm]{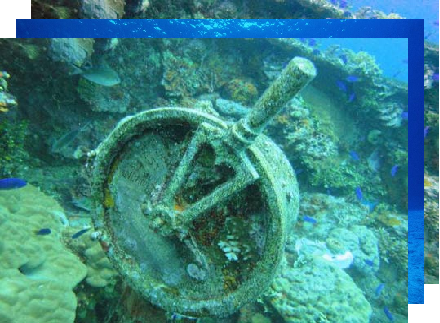}}
\subfigure [Wreck] {
\includegraphics[height=2cm,width=2.5cm]{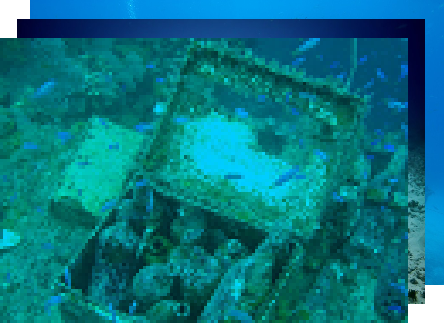}}
\subfigure [Diver] {
\includegraphics[height=2cm,width=2.5cm]{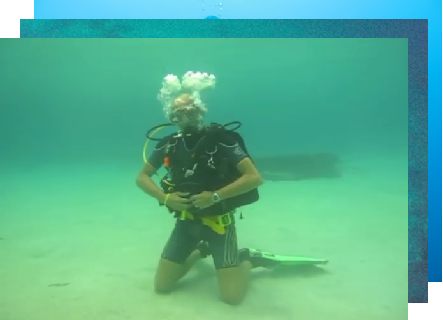}}\\
\vspace{-0.3cm}
\caption{\small  Typical underwater images in UIED, including 16 coral images, 26 marine life images, 14 seabed rock images, 12 sculpture images, 10 wreck images and 22 diver images.}
\label{fig:7}
\end{figure}

We select 100 authentic underwater images from Google and related works \cite{Sun2010}, \cite{Li2020}. These underwater images are taken from real underwater scenes, with resolutions ranged from 183 $\times$ 275 to 1350 $\times$ 1800. Typical underwater images are presented in Fig. \ref{fig:7}. To generate more enhanced images for test, we utilize 10 representative UIE algorithms, including 3 non-physical-model-based methods (\emph{i.e.} fusion-based \cite{Ancuti2012}, retinex-based \cite{Fu2014}, two-step-based \cite{Fu2017}), 4 physical-model-based methods (\emph{i.e.} histogram prior \cite{odm2016}, UDCP \cite{Drews2016}, UIBLA \cite{Peng2017}, GDCP \cite{Peng2018}), 2 deep-learning-based methods (\emph{i.e.} UWCNN \cite{Li2019}, Water-Net \cite{Li2020} and 1 commercial application (\emph{i.e.} dive+ \cite{dive+}). With the 100 images and 10 UIE approaches, we have a total of 1,000 enhanced underwater images. All enhancement underwater images and the corresponding raw images are included in the UIED.

%

\subsection{Subjective Test}
We implement a subjective quality study to evaluate the 1,000 enhanced  images. In the subjective quality evaluation, we adopt a single-stimulus strategy and a five-level quality scale to label images \cite{Recommendation2012}. A lower rating score represents a worse perceptual visual quality, and vice versa. A detailed description of the rating criteria is given in Table \ref{tab:1}.

The subjective test consists of 10 sessions and in each session, 100 of the 1,000 images are evaluated. At the beginning of each session, a subject watches and evaluates 5 prescreen images to get familiar to testing environment and procedure. Then, the subject is asked to score the following 100 images in a random order. Each image is shown for 7 seconds before next. The subject can take a 5-minute break between sessions to avoid visual fatigue.

There are totally 16 subjects including 9 males and 7 females in the subjective test. They all have prior knowledges of image processing. All test images are shown in random order with a MATLAB graphical user interface. The images are displayed in laboratory environment with normal illumination conditions. All of the above conditions are set and calibrated according to the recommendations of ITU-R  \cite{Recommendation2012}. Table \ref{tab:2} lists an overview of the test methodology and conditions.

\begin{table}
\caption{Rating Criteria for Subjective Test}

  \centering
  \begin{tabular}{c|c}
    \toprule
    Level & Description\\
    \noalign{\global\arrayrulewidth1pt}\hline\noalign{\global\arrayrulewidth0.01pt}   
    1 & The target is invisible, and color severely distorts.\\  
    2 & The target is invisible, and color partially distorts.\\  
    3 & The target is visible, and color slightly distorts.\\  
    4 &	The target is visible, some flaws in visual quality.\\
    5 & The target is clearly, and visual quality is great.\\
    \bottomrule
  \end{tabular}

  \label{tab:1}
\end{table}

\begin{table}[h]

\caption{Detailed Setting of Subjective Experiment}

\label{tab:2}
\setlength{\tabcolsep}{1.2mm}
\begin{tabular}{ccc}
\toprule
Category 	& Item   &  Detail   \\  \midrule
   \multirow{3}{*}{Display}	& Monitor        	& AOC 24n2h LCD           		\\
	& Resolution		& 1920 × 1080				\\
	& Platform 		& Matlab R2016b				\\ 
	\hline
	\multirow{3}{*}{Methodology}
	& Method & Single-stimulus\\
    
    & Quality scale & 5-level categorical\\
    
    & Order & Random\\
    \hline
   \multirow{4}{*}{Test settings}
   & Groups & 10\\
    
   & Subjects number& 9 males / 7 females\\
   & Time interval & 7 seconds\\
   & Environment & Laboratory\\
\bottomrule   
\vspace{-0.5cm} 
\end{tabular}
\end{table}

\subsection{Data Post-Processing}
To show the reliability of data, we choose to use Normalized Cross Correlation (NCC) and Euclidean distance (EUD) to evaluate the agreement of subject ratings \cite{Ma2012}. A higher NCC value or a lower EUD indicates higher correlation between two subjective rating vectors. For the above 16 subjective ratings, the average value of NCC and EUD is 0.961 and 0.055, respectively. Therefore, the subject ratings are agreed on the perceptual qualities of these images and the testing results are reliable.

%


\begin{figure}[]
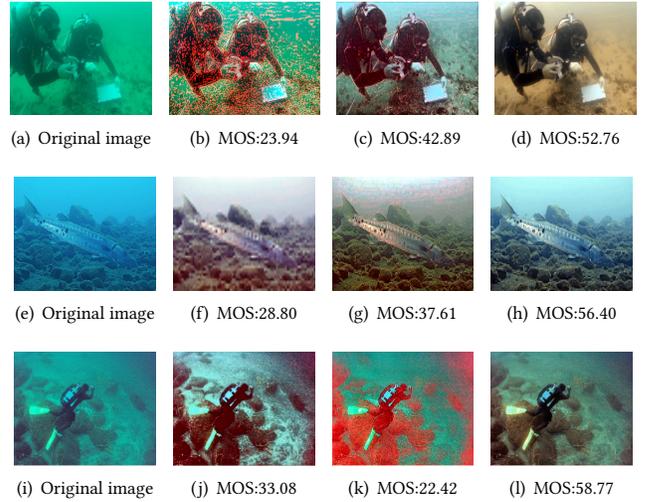

\centering
	\subfigure [Original image]{
	\includegraphics[height=1.5cm,width=1.875cm]{./e_1.png}	}
	\subfigure [MOS:23.94]{
	\includegraphics[height=1.5cm,width=2cm]{./e_2.png}	}
	\subfigure [MOS:42.89]{
	\includegraphics[height=1.5cm,width=1.875cm]{./e_3.png}	}
	\subfigure [MOS:52.76]{
	\includegraphics[height=1.5cm,width=1.875cm]{./e_4.png}	}\\
	\subfigure [Original image]{
	\includegraphics[height=1.5cm,width=1.875cm]{./f_1.png}	}
	\subfigure [MOS:28.80]{
	\includegraphics[height=1.5cm,width=1.875cm]{./f_2.png}	}
	\subfigure [MOS:37.61]{
	\includegraphics[height=1.5cm,width=1.875cm]{./f_3.png}	}
	\subfigure [MOS:56.40]{
	\includegraphics[height=1.5cm,width=1.875cm]{./f_4.png}	}\\
	\subfigure [Original image]{
	\includegraphics[height=1.5cm,width=1.875cm]{./g_1.png}	}
	\subfigure [MOS:33.08]{
	\includegraphics[height=1.5cm,width=1.875cm]{./g_2.png}	}
	\subfigure [MOS:22.42]{
	\includegraphics[height=1.5cm,width=1.875cm]{./g_3.png}	}
	\subfigure [MOS:58.77]{
	\includegraphics[height=1.5cm,width=1.875cm]{./g_4.png}	}\\
	
	\caption{\small The MOS values of images in Fig. 1.}
	\label{fig:9}
\end{figure}

We then follow the steps in \cite{Min2017} to process the subjective ratings. Rating for an image is considered as outlier if it is outside 2 or $\sqrt{20}$  standard deviations of the mean rating of that image. A subject with more than 5\% outlier evaluations is rejected. Both outlier ratings and outlier subjects are excluded from the following processing. The normalized ratings for an image are averaged over all valid subjects to the Mean Opinion Score (MOS). The MOS of the database is almost ranged from 20 to 80. That is, our collected subjective scores span a wide range from low to high scores.

Fig.\ref{fig:9} shows the MOS values corresponding to the enhancement results in Fig.\ref{fig:1}. It can be seen that our subjective experiment rated has a high correlation to HVS. On the other hand, the popular UCIQE and UIQM show inferior performances that still have room to be further improved.

\section{Proposed UIF Metric}
Based on the UIED, we are able to develop the UIF metric for a more accurate quantitative evaluation of enhanced underwater images. The design philosophy lies in three key aspects, including naturalness, sharpness, and structure of an enhanced underwater image. In this section, we present to attract features to characterize the naturalness, sharpness and structure, and further integrate them to infer the ultimate quality score. Details of our method are presented as follows.

\subsection{Naturalness-Related Features}
In most cases, the natural underwater images have low color richness due to light absorption and scattering. Thus, many UIE methods aim to eliminate the color attenuation from vision and improve color richness. However, overemphasis on color may also result in unreasonable color or an unnatural look.  Therefore, we employ the NSS regularities to approximately capture the attributes of naturalness of underwater images. In addition, researches have shown a good correlation between underwater colors and human perception in CIELab space \cite{yang2015}. Inspired by this, we also transfer the underwater images to CIELab space to calculate the naturalness-related features.

\emph{Distribution of NSS model.} To detect naturalness-related features in enhanced underwater images, we first choose the NSS model which is frequently used in IQA tasks \cite{Yang2016, Yang2017}. It is defined on the brightness of image



\begin{equation}
  \begin{split}
  f(x;\nu,\sigma^2)=\dfrac{\nu}{2\delta\Gamma(\dfrac{1}{\theta})}e^{(-(\dfrac{|x|}{\theta})^\nu)}, 	
  \end{split}\	
\label{func:10}
\end{equation}

\noindent where $\delta$ and $\Gamma(\cdot)$ are defined as:
\begin{equation}
  \begin{split}
  \theta=\sigma\sqrt{\dfrac{\Gamma(1/ \nu)}{\Gamma(3/ \nu)}},
  \end{split}\	
\label{func:11}
\end{equation}

\begin{equation}
  \begin{split}
  \Gamma(\theta)=\int_{0}^{\infty} t^{\theta-1}e^{-t} dt  \quad \theta>0.
  \end{split}\	
\label{func:12}
\end{equation}

\noindent The parameter $\nu$ and $\sigma^2$ control the shape and variance of the distribution, respectively, which are collected to describe image naturalness.



\emph{Contrast and variance in CIELab.}  We transform the underwater image from RGB to CIELab, which is a uniform color space. Considering the luminance contrast is one of the most sensitive factors in poor illumination, we calculate the luminance contrast for enhanced images in turbid underwater environment. It is obtained by the ratio between highest and lowest luminance values


%

\begin{equation}
  \begin{split}
  \sigma_{\rm CIE}=\dfrac{1}{N} \sum_{i=1}^N \dfrac{\sqrt{\alpha_i^2+\beta_i^2}}{L_i},
  \end{split}\	
\label{func:12}
\end{equation}
\noindent where $\alpha$ and $\beta$ are the channel parameters of CIELab space, $L$ is value in luminance channel and $N$ presents the number of image pixels.


In summary, the naturalness-related features include 
\begin{equation}
  \begin{split}
 \emph F_{\rm  naturalness}=\lbrace\nu,\sigma^2, C_{\rm CIE}, \sigma_{\rm CIE} \rbrace.
  \end{split}\	
\label{func:12}
\end{equation}
\subsection{Sharpness-Related Features}
A high-fidelity image is usually with rich details that are characterized by sharpness of edges and pixels. In underwater imaging, forward scattering may affect image details and blurs edges. As a result, an important objective of UIE is to improve the sharpness of edges and thus present high-quality pictures with clear objects. To characterize the sharpness of an enhanced image, we exploit the following features including Dark Channel Prior (DCP) index, contrast, edge contrast and entropy.

\emph{DCP Index.} The index was utilized to capture image contrast during haze removal \cite{2011Single}. In underwater environment with low light intensities, dark images are usually captured, which might hinder object recognition by humans or algorithms. Therefore, the UIE approaches are designed to enhance the lightness of underwater images, which changes the DCP channel pixels. Inspired by this, we compute the average value of DCP channel pixels as:
\begin{equation}
  \begin{split}
  \mu_{\rm dark}= {\rm average}\lbrace{\underset{c \in {R,G,B}}\min I_c\rbrace},
  \end{split}\	
\label{func:13}
\end{equation}
where $c \in R, G, B$ indicates the RGB channels of enhanced image $I$.

\emph{Contrast.} The contrast has been utilized in visual enhancement of underwater images \cite{odm2016}. First of all, the enhanced image is divided into non-overlapped grayscale patches with size $64\times64$. Then, a patch is labeled as a textured patch if its edge density, {\it i.e.} the ratio between edge pixels and all pixels within the patch, is larger than 0.2\%. Finally, the overall contrast is obtained as the sum of standard variances of all textured patches:

\begin{equation}
  \begin{split}
  C=\sum_{i=1}^M {\rm std\_var}(P_{i,j}),
  \end{split}\	
\label{func:13}
\end{equation}

\noindent where $P_{i,j}$ is the $i,j$-th patch of image and $M$ represents the number of textured patches.

\emph{Edge contrast.} The edge information has been widely utilized to assess fidelities of images \cite{Sharp2, Sharp3}. In this work, we utilize a simple but efficient edge extraction for ease of calculation.  First of all, the Canny edge detector is applied to all channels of image to obtain three edge maps. Then, each edge map is equally divided into $m\times n$ non-overlapped blocks with size 5$\times$5. Finally, the contrasts of all blocks are calculated and averaged to obtain a final measure within edge maps \cite{2014Choosing}:

\begin{equation}
  \begin{split}
  C_{\rm edge}=\sum_{c=1}^3\lambda_c\cdot\ \dfrac{2}{m n} \sum_{i=1}^{m}\sum_{j=1}^{n} \log(\dfrac{ {\rm max}B_{c,i,j}}{{\rm min} B_{c,i,j}}),
  \end{split}\	
\label{func:13}
\end{equation}

\noindent where $B_{c,i,j}$ represents the $i,j$-th edge block in channel $c$, and $\max$ and $\min$ are to calculate the extreme values within the edge block. $\lambda_c$ is a coefficient for RGB color channels. 

\emph{Entropy.} As a classic image measurement, the entropy characterizes diversity of image pixels. Therefore, it has a high correlation to image contrast. For the enhanced image with $L$ brightness levels, the entropy is defined by

\begin{equation}
  \begin{split}
  E=-\sum_{i=0}^{L-1}p_i \log p_i,
  \end{split}\	
\label{func:13}
\end{equation}

\noindent where $p_i$ is the histogram probability of brightness value $i$. 

In summary, the sharpness-related features include

\begin{equation}
  \begin{split}
  F_{\rm sharpness}=\lbrace{\mu_{\rm dark}, C, C_{\rm edge}, E }\rbrace.
  \end{split}\	
\label{func:13}
\end{equation}

\begin{figure}[b]
\centering
	\subfigure[] {
	\includegraphics[height=2.7cm,width=4cm]{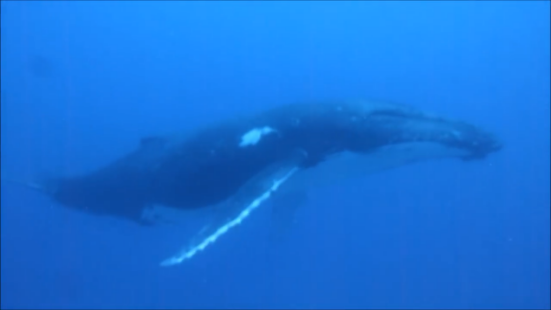}	}
	\subfigure []{
	\includegraphics[height=2.7cm,width=4cm]{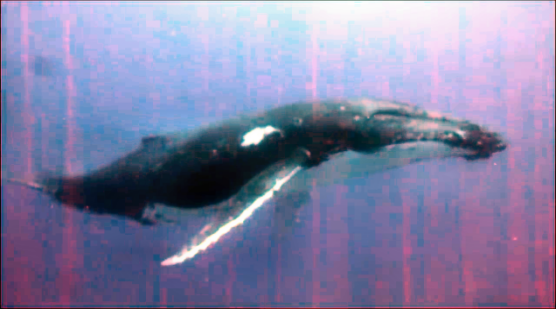}	}
	\vspace{-0.2cm}
	\caption{An example of structural artifact introduced by UIE. }
	\label{fig:6}
\end{figure}

\begin{figure}[b]
\small
\centering
	\subfigure [original image]{
	\includegraphics[height=2cm,width=2.5cm]{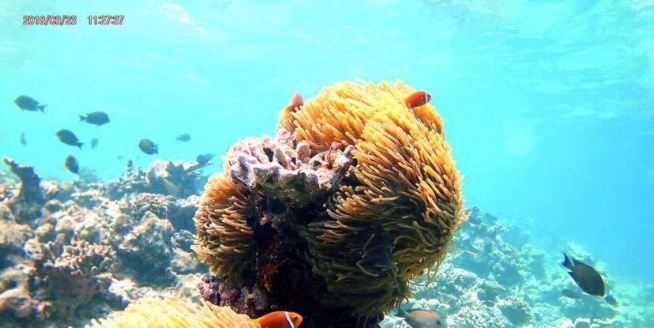}	}
	\subfigure [over-enhanced image]{
	\includegraphics[height=2cm,width=2.5cm]{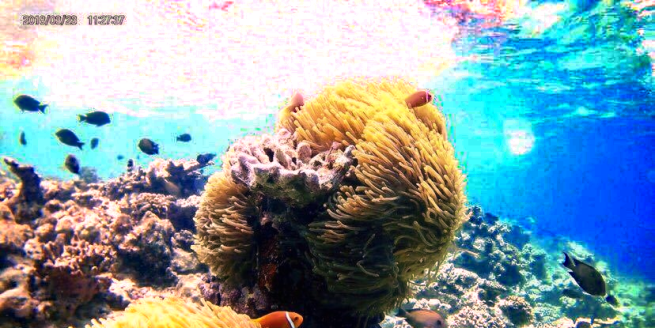}	}
	\subfigure [variance similarity map]{
	\includegraphics[height=2cm,width=2.5cm]{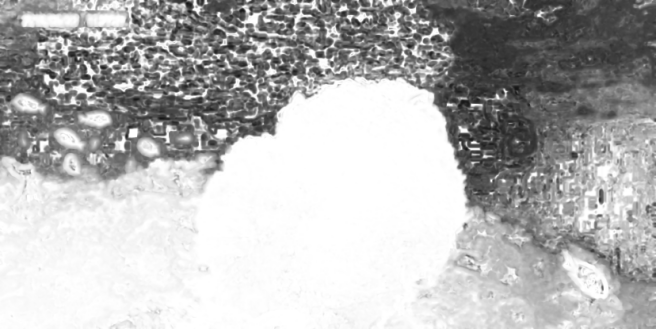}	}
	\vspace{-0.2cm}
	\caption{\small An example of over-enhanced underwater images
and variance similarity map. (a) original image, (b) over-enhanced image, (c) variance similarity map.}
	\label{fig:13}
\end{figure}

\begin{figure*}[htbp]
\hspace{5cm}
\includegraphics[height=10.0cm,width=16.0cm]{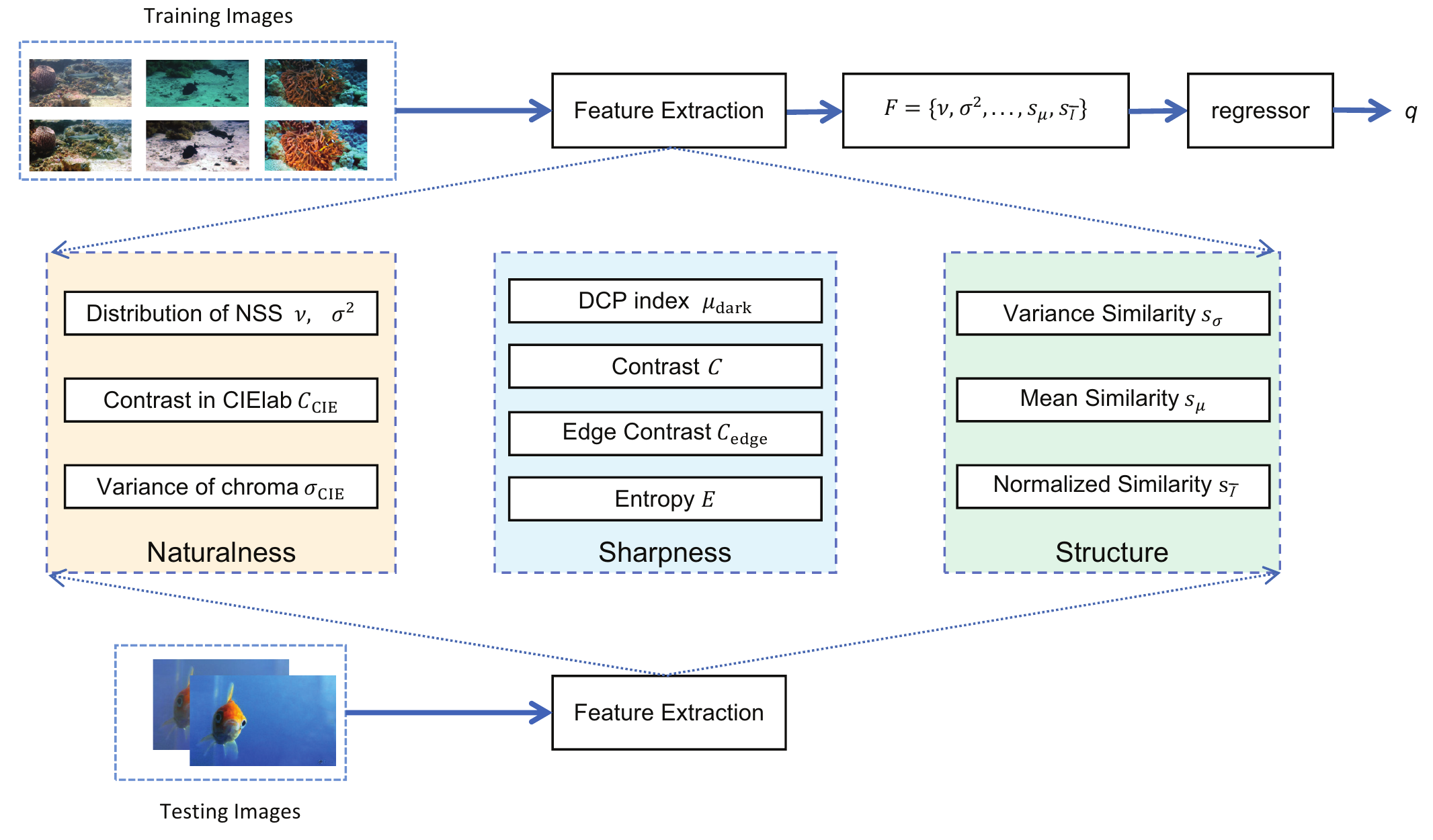}
\caption{The detailed framework of the proposed UIF method.}
\label{fig:11}
\end{figure*}

\subsection{Structure-Related Features}

The structural similarity between original and enhanced images is critical because the UIE algorithms, especially the deep-learning-based UIE, may introduce structural artifacts during the enhancement process. An example is shown in Fig. \ref{fig:6}, where unwanted textures are introduced due to color changes. These artifacts may severely degrade visual quality of enhanced images. Therefore, we need to measure the structural similarity between original and enhanced images. It is noted that the image structural similarity was firstly introduced by SSIM. To evaluate the similarity between an image and its distorted version. Inspired by this index, we calculate our structure-related features with brightness values of original and enhanced underwater images. 

\emph{Variance similarity.} In most cases, over enhancement of underwater images will cause large variance of brightness, with an example shown in Fig. \ref{fig:13} (b). In such case, we can construct a variance similarity measure to identify these over-enhanced regions. Similar to SSIM, we define the variance similarity as:
\begin{equation}
  \begin{split}
  s_{\sigma}=\dfrac{2\sigma_I\sigma_0 +c_1}{\sigma_I^2+\sigma_0^2+c_1},
  \end{split}\
\label{func:13}
\end{equation}

\noindent where $\sigma_I$ and $\sigma_0$ are the variances of enhanced and original images, respectively. They are calculated locally with a 7$\times$7 window. $c_1$ is a small constant to avoid zero denominators. With this measure, we can calculate the similarity between Fig. \ref{fig:13} (a) and (b) to obtain  the variance map shown in Fig. \ref{fig:13} (c). It can be readily seen that the over-enhanced regions with large brightness changes are labeled with low variance similarities. 
 

\emph{Mean and normalized similarities.} 
The local mean $\mu$ and normalized image $\bar I$ can describe the perceptual similarity of textured regions \cite{Gu2017}, where the normalized image is calculated by $(I-\mu(I))/\sigma(I)$. Therefore, we also utilize the similarities between mean and normalized images as supplementary indexes:
\begin{equation}
  \begin{split}
  s_{\mu}=\dfrac{2\mu_I\mu_0 +c_2}{\mu_I^2+\mu_0^2+c_2},
  \end{split}\	
\label{func:13}
\end{equation}

\begin{equation}
  \begin{split}
  s_{\bar{I}}=\dfrac{2\bar I_I \bar I_0 +c_3}{\bar I_I^2+\bar I_0^2+c_3},
  \end{split}\	
\label{func:14}
\end{equation}

\noindent where $\mu_I$ and $\mu_0$ respectively denote the local mean values of enhanced and original images, $\bar I_I$ and $\bar I_0$ respectively denote the normalizations of enhanced and original images, $c_2$ and $c_3$ are small constants to avoid zero denominators.
 
In summary, the structure-related features include
\begin{equation}
  \begin{split}
  F_{\rm structure}=\lbrace{s_{\sigma},s_{\mu},s_{\overline{I}}}\rbrace.
  \end{split}\	
\label{func:14}
\end{equation}

\subsection{Feature Pooling and Regression}
Fig. \ref{fig:11} shows the detailed framework of proposed UIF metric, which mainly consists of two modules: feature extraction and feature regression. In feature extraction module, we obtain the naturalness-related, sharpness-related and structure-related features, as discussed above. If a feature is represented by a two-dimensional map, an average pooling is employed to reduce the dimensions. All features are then linked into a feature vector:
\begin{equation}
  \begin{split}
F = [ \nu, \sigma^2, C_{\rm CIE}, ... ,s_{\sigma}, s_{\mu},s_{\bar{I}}].
\end{split}\ 
\label{func:16}
\end{equation}

In feature regression module, we select the SVR for fusion considering its success in regression tasks. As illustrated in Fig. \ref{fig:11}, we use labeled underwater enhanced image pairs to train the regressor, which can be utilized to predict the quality of any input image. Given the features $F = [ \nu, \sigma^2, C_{\rm CIE}, ... ,s_{\sigma}, s_{\mu},s_{\bar{I}}]$, the corresponding quality label MOS values $S_i$ and the training image set, we can train the regressor by SVR:
\begin{equation}
  \begin{split}
{\rm regressor}={\rm TRAIN}(F_i, S_i).
\end{split}\ 
\label{func:17}
\end{equation}
After the training process, we can apply this regressor to yield quality scores of any  testing enhanced image $I$ and original image $I_0$ :
\begin{equation}
  \begin{split}
q={\rm PREDICT}(I, I_0,\rm regressor).
\end{split}\ 
\label{func:18}
\end{equation}
LIBSVM \cite{2007LIBSVM} is adopted to implement SVR and has previously been applied to IQA problems\cite{fang2015}. In training process, we choosea Radial Basis Function (RBF) kernel. The other SVR parameters are set as: penalty coefficient=0.1, $\varepsilon$-insensitive loss function=0.01, and RBF kernel parameters=1.

\section{EXPERIMENTAL RESULTS}
In this section, we will evaluate proposed UIF metric with experimental results. Popular IQA metrics, including general no-reference metrics, enhanced image quality metrics and UIE quality metrics are examined for comparison. We also present ablation study to show the effectiveness of all types of features.

\subsection{Experiment Settings }

To demonstrate the efficiency of our method, we choose a variety of IQA methods for comparison. They include 9 no-reference IQA methods for natural images (BRISQUE \cite{Anish2012}, NIQE \cite{Anish2013}, NFERM \cite{Gu2015}, IL-NIQE\cite{Zhang2015}, SISBLIM \cite{GU2014}, BLIINDS-II \cite{Xue2014}, dipIq \cite{dipiq}, og-iqa \cite{ogiqa}, CNN-IQA \cite{cnniqa}), 3 IQA methods for enhanced images (CPCQI \cite{Gu2018}, BIQME \cite{Gu2018}, and NR-CDIQA \cite{fang2015}) and 2 IQA methods for enhanced underwater images (UCIQE \cite{yang2015}  and UIQM \cite{K2016}). For fair comparison, we use the publicly available codes provided by authors. If a metric is obtained by machine learning and its training code is available, its parameters are further tuned for fair comparison. To examine the generalization ability of machine learning models including our UIF metric, they are examined with $k$-fold cross-validations, where $k$ is set as 4. A fold of data is examined and recorded only when it is used as testing test. 
 
The performances of all IQA metrics are evaluated by two commonly used consistency criteria, including Spearman Rankorder Correlation Coefficient (SRCC) and Pearson Linear Correlation Coefficient (PLCC).  The SRCC scores the prediction monotonousness to MOS values, while the PLCC scores the linear correlation between the IQA model’s predictions and MOS values. In particular, higher PLCC or SRCC indicates an IQA metric is more consistent with subjective quality evaluations.

\subsection{Comparisons and Discussions}

The comparison results of all IQA metrics are summarized in Table \ref{tab:4}, where the best results and 2nd-best results are highlighted with red bold and blue bold, respectively. As shown in the table, some no-reference IQA methods are designed for images in the air (\emph{e.g.} BRISQUE, NFERM, NIQE, IL-NIQE, OG-IQA, SISBLIM, and BLIINDS-II), thus their extracted NSS characteristics are not applicable to all underwater images. The accuracy of these no-reference IQA metrics is low. Even for learning-based metrics such as dipIQ, CNN IQA, the performance in terms of SRCC and PLCC are not high. This fact implies the particularity and complexity of underwater image characteristics, which makes the IQA for UIE images a more challenging problem.

\begin{table}
\caption{Performance Comparison of Selected IQA metrics}
\label{tab:4}
\setlength{\tabcolsep}{3.5mm}
 \begin{tabular}{cccc}
\toprule
Type & Methods & SRCC & PLCC  \\
\midrule
 \multirow{9}{*}{General IQA}&BRISQUE& {\textcolor{blue} {\textbf {0.465}}}&{\textcolor{blue} {\textbf {0.496}}}\\
&NFERM&0.355&0.339\\
&NIQE&0.326&0.344\\
&IL-NIQE&0.393&0.347\\
&OG-IQA&0.216&0.233\\
&SISBLIM&0.277&0.321\\
&BLIINDS-II&0.341&0.352\\
&dipIQ&0.126&0.217\\
&CNN IQA&0.027&0.081\\
\midrule
\multirow{3}{*}{IQA for enhancement}
&CPCQI&0.266&0.284\\ 
&BIQME&0.205&0.259\\
&CDIQA&0.276&0.292\\
\midrule
\multirow{3}{*}{IQA for UIE}
&UCIQE&0.252&0.298\\
&UIQM&0.276&0.268\\

&UIF&{\textcolor{red} {\textbf {0.733}}}&{\textcolor{red} {\textbf {0.757}}}\\
\bottomrule
\end{tabular}
\end{table}

%
The methods designed for enhanced images (\emph{e.g.} CPCQI, BIQME, and CDIQA) also show inferior performances in UIE images. This may be because they are designed to prefer images with high contrasts without consideration to the special distortions and artifacts in underwater images. As a result, the features extracted by these methods are not competitive and the experimental results are not ideal.

Besides, the performances of popular UCIQE and UIQM are still low in UIED. The possible reasons are as follows. Firstly, these methods were proposed earlier when few UIE algorithms were developed. The recent booming of UIE approaches have led to a variety of image distortions and artifacts, which cannot be considered by early methods. Secondly, they were tuned in small datasets due to lack of large-scale dataset, which could not be constructed due to lack of UIE algorithms. Thirdly, they incorporated fewer features to evaluate underwater images, which limits their performance in diverse underwater scenarios. 

Finally, the proposed UIF metric achieves superior performance to other metrics. This demonstrates the effectiveness of multiple features and SVR-based fusion. More importantly, our metric is also benefited from the large-scale UIED, which is the largest-ever underwater image quality database with subjective scores.

\begin{figure}[h]
\fboxsep=0.1mm 
\fboxrule=0.1pt 
\vspace{-0.25cm}
\hspace{-0.5cm}
\small
	\centering

	\subfigure [0.544/41.59]{
	\includegraphics[width=1.9cm,height=1.5cm]{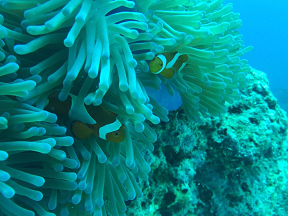}	}
	\subfigure [0.714/38.63]{
	\includegraphics[width=1.9cm,height=1.5cm]{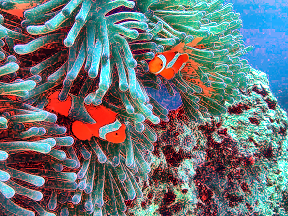}	}
	\subfigure [0.643/55.54]{
	\includegraphics[width=1.9cm,height=1.5cm]{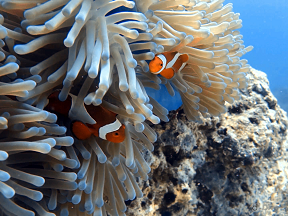}	}
	\subfigure [0.627/28.37] {
	\includegraphics[width=1.9cm,height=1.5cm]{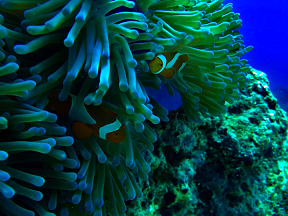}	}\\
	\vspace{-0.25cm}
	\subfigure [0.465/34.10]{
	\includegraphics[width=1.9cm,height=1.5cm]{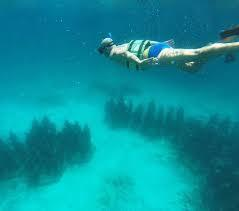}	}
	\subfigure [0.696/29.51]{
	\includegraphics[width=1.9cm,height=1.5cm]{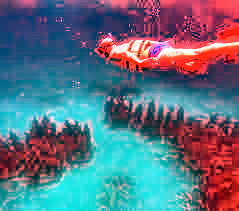}	}
	\subfigure [0.591/42.21] {
	\includegraphics[width=1.9cm,height=1.5cm]{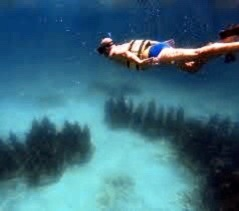}	}
	\subfigure [0.631/43.09]{
	\includegraphics[width=1.9cm,height=1.5cm]{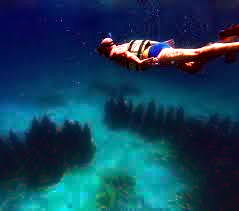}	}\\
	\vspace{-0.25cm}
	\subfigure [0.640/55.54] {
	\includegraphics[width=1.9cm,height=1.5cm]{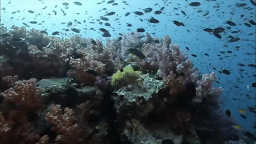}	}
	\subfigure [0.644/73.81]{
	\includegraphics[width=1.9cm,height=1.5cm]{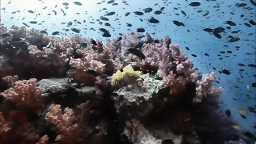}	}
	\subfigure [0.643/72.60] {
	\includegraphics[width=1.9cm,height=1.5cm]{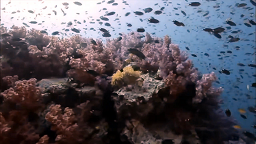}	}
	\subfigure [0.674/54.54]{
	\includegraphics[width=1.9cm,height=1.5cm]{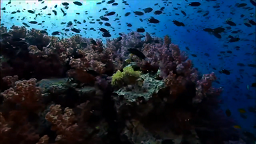}	}\\
	\vspace{-0.25cm}

	\subfigure [0.432/28.37]{
	\includegraphics[width=1.9cm,height=1.5cm]{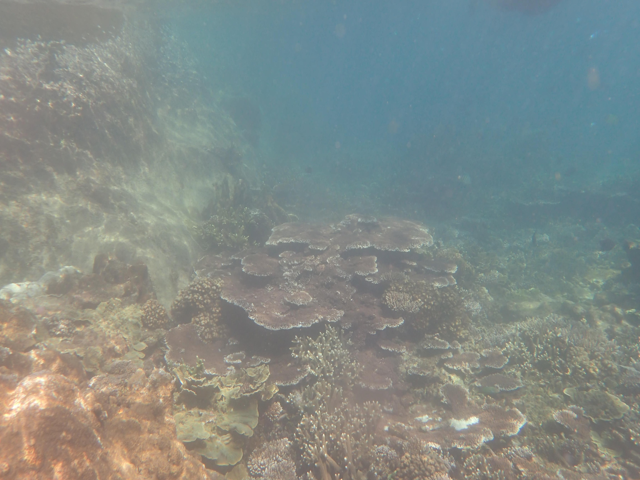}	}
	\subfigure [0.654/58.25]{
	\includegraphics[width=1.9cm,height=1.5cm]{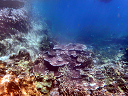}	}
	\subfigure [0.593/52.37]{
	\includegraphics[width=1.9cm,height=1.5cm]{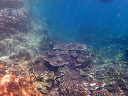}	}
	\subfigure [0.751/54.97]{
	\includegraphics[width=1.9cm,height=1.5cm]{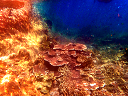}	}\\
    \vspace{-0.1cm}
	\caption{\small Typical underwater images (the first left column) and their scored enhancements (the other three columns), where the scores of each enhanced image are respectively given by UCIQE and proposed UIF. }
	
	\label{fig:12}
\end{figure}

%
To present an intuitive comparison between UIF and conventional metrics, we select four typical underwater images and their enhancements, as shown in Fig. \ref{fig:12}. We also calculate the UCIQE and UIF values of all enhanced images and present them in the same figure. From this figure, the conventional UCIQE metric may focus on chroma components of images and thus tend to prefer images with rich colors. It gives high scores to the (b) (f) (l) (p), which have either red artifacts or contrast distortions. In contrast, the UIF metric has preferences for (c) (h) (j) (n). This is because the UIF considers naturalness, contrast and similarity besides of underwater image colors, which are more consistent with HVS. Therefore, our UIF metric shows the state-of-the-art performance in quality evaluation of enhanced underwater images.

\subsection{Ablation Study}

To evaluate the contribution of each type of features, we conduct a series of ablation experiments. Specifically, the image features merged in the following methods.
 
\noindent \textbf{•} Method1: Only naturalness-related features.\\
\textbf{•} Method2: Only sharpness-related features.\\
\textbf{•} Method3: Only structure-related features.\\
\textbf{•} Method4: Naturalness-related and sharpness related features.\\
\textbf{•} Method5: Sharpness-related and structure-related features.\\
\textbf{•} Method6: Naturalness-related and structure-related features.\\
\textbf{•} Method7: The UIF with all three types of features.\\

Table \ref{tab:11} shows the performances of all 7 methods. From the table, we can draw several conclusions. Firstly, the model achieves acceptable performances even with one type of feature. Compared with Table \ref{tab:4}, the corresponding performance is still superior to other IQA models in underwater IQA. Secondly, with more types of features, the proposed model generally improves its prediction performance in terms of SRCC and PLCC. Thirdly, by a fusion of all types of features, our UIF metric achieves the best performance of all. This fact demonstrates the effects of all types of features in our UIF metric. With an SVR-based regression, the fused index achieves superior performance to predict subjective evaluations with an objective approach.

\begin{table}[h]
\caption{Performance of Different Feature Groups}

\label{tab:11}
\setlength{\tabcolsep}{1.5mm}
\begin{tabular}{cccc}
\toprule
Feature Groups 	& Descriptions   &  SRCC  & PLCC \\  \midrule
Method 1	& Only naturalness         	& 0.663      	& 0.689     		\\
Method 2	& Only sharpness 		& 0.588      	& 0.605				\\
Method 3	& Only structure  		& 0.618			& 0.645				\\ 
Method 4	& Naturalness and sharpness 		& 0.676			& 0.691				\\ 
Method 5	& Sharpness and structure  		& 0.707			& 0.713				\\ 
Method 6	& Naturalness and structure  		& 0.701	& 0.727\\
Method 7	& All type features  	& \textbf{0.733}	& \textbf{0.757}\\\bottomrule   
\end{tabular}
\end{table}

\subsection{Further Applications}
The proposed UIF metric can be further applied in other IQA tasks of enhanced images, with acceptable correlations to subjective scores. In this section, we test the UIF metric in three other databases, including DHQ \cite{Yeganeh2015} for dehazing quality assessment, CID2013\cite{CID2013} and CCID2014 \cite{CCID2014} for contrast-enhanced quality assessments. For comparison, we keep the same SVR parameter settings and training steps. The training to testing ratio is set as 80:20. In Table 5, we present average SRCC and PLCC values for the three databases. Although the existing IQA methods for general enhanced images cannot work well in underwater IQA, as shown in Table 3, our UIF metric can still achieve acceptable performances in for IQA of general enhanced images. This also shows the generalization of our UIF metric to a certain extent. 

\begin{table}[h]
\caption{Evaluations on Other IQA Tasks}

\label{tab:7}
\setlength{\tabcolsep}{2mm}
\begin{tabular}{cccc}
\toprule
Databases &Type	&  SRCC	& PLCC  \\  \midrule
CID2013	& Contrast-changed Image	&0.6959	      	&0.6795	      	\\
CCID2014	& Contrast-changed Image & 0.7167 	& 0.7031  	\\
DHQ		& Dehazing Image 	& 0.6234     & 0.6707	\\ 
  \bottomrule    
\end{tabular}
\end{table}


\section{CONCLUSIONS}
Recent efforts of UIE have greatly promoted the visual quality of underwater images. Nevertheless, the quality measures of UIE have not been updated to evaluate the enhanced images by newly developed UIE models. In this paper, we make the first attempt to build a large-scale underwater image quality database for UIE, namely UIED, which can be utilized to train and evaluate objective IQA approaches for enhanced underwater images. Based on this database, we also propose a new quality metric of underwater images, namely UIF, which shows the state-of-the-art performance in this scenario. Experimental results also demonstrate the effectiveness of features and the generalization ability of UIF.  The proposed metric can evaluate the qualities of enhanced underwater images and also help to select optimal UIE approaches under different underwater environments.

\begin{acks}

\end{acks}

\bibliographystyle{ACM-Reference-Format}

\bibliography{ref1}

\end{document}